\documentclass[11pt,halfline,a4paper]{ouparticle}

\usepackage{setspace}

\usepackage{amsthm} 
\theoremstyle{definition}
\newtheorem{definition}{Definition}[section] 
\newtheorem{theorem}{Theorem}[section]

\usepackage{graphicx}
\usepackage{subcaption} 
\usepackage{float}      

\usepackage{multirow}
\usepackage{booktabs} 
\usepackage{caption}
\usepackage{xcolor} 
\usepackage{array}

\usepackage{amsmath}   
\usepackage{amssymb}   
\usepackage{amsfonts}

\usepackage{mathrsfs}  
\usepackage{bbm}       
\usepackage{bm}

\usepackage{algorithm, algpseudocode, float}

\usepackage{titlesec}
\titleformat{\section}[block]{\large\bfseries}{\thesection.}{0.6em}{}
\titleformat{\subsection}[block]{\normalsize\bfseries}{\thesubsection.}{0.3em}{}

\begin{document}

	\title{Hyperspectral Image Spectral-Spatial Feature Extraction via Tensor Principal Component Analysis}
	
	\author{
		{\small{}Yuemei Ren, Liang Liao, Stephen John Maybank, Yanning Zhang, and Xin Liu}		
		\thanks{Corresponding author: Liang Liao (liaoliang@ieee.org or liaoliangis@126.com). 
			L. Liao, S. J. Maybank, and Y. Ren contributed equally to this work.
			L. Liao is with the School of Electronics and Information, Zhongyuan University of Technology, Zhengzhou 450007, Henan, China.
			S. J. Maybank (sjmaybank@dcs.bbk.ac.uk) is with Birkbeck College, University of London, London, UK, WC1E 7HX.
			Y. Ren is with the Department of Electronic Information Engineering, Henan Polytechnic Institute, Nanyang 473000, China.
			Y. Zhang is with the School of Computer Science, Northwestern Polytechnical University, Xi'an 710072, China.
			X. Liu is with the Information Center of the Yellow River Conservancy Commission, Zhengzhou 450000, China.
			L. Liao and Y. Ren are also affiliated with the School of Computer Science, Northwestern Polytechnical University, Xi'an 710072, China.
		}
	}

	\abstract{
		This paper addresses the challenge of spectral-spatial feature extraction for hyperspectral image classification by introducing a novel tensor-based framework. The proposed approach incorporates circular convolution into a tensor structure to effectively capture and integrate both spectral and spatial information. Building upon this framework, the traditional Principal Component Analysis (PCA) technique is extended to its tensor-based counterpart, referred to as Tensor Principal Component Analysis (TPCA). The proposed TPCA method leverages the inherent multi-dimensional structure of hyperspectral data, thereby enabling more effective feature representation. Experimental results on benchmark hyperspectral datasets demonstrate that classification models using TPCA features consistently outperform those using traditional PCA and other state-of-the-art techniques. These findings highlight the potential of the tensor-based framework in advancing hyperspectral image analysis.}

	\date{\footnotesize{}Published in IEEE Geoscience and Remote Sensing Letters, September 2017}


	\maketitle

	\section{Introduction}
	\label{sec1}

	Hyperspectral imaging sensors collect hyperspectral images as three-dimensional (3D) arrays, with two spatial dimensions representing the image width and height, and a spectral dimension corresponding to the spectral bands—often numbering over one hundred. Given the redundancy in the raw data representation, it is advantageous to design effective feature extractors that leverage the spectral information of hyperspectral images \cite{1,2}. For instance, Principal Component Analysis (PCA) reduces high-dimensional data to lower-dimensional feature vectors through linear projections derived from eigenanalysis. By selecting the principal components, the resulting feature vectors retain most of the available information.

	However, PCA, like other vector-based methods, lacks a built-in mechanism to capture the spatial information inherent in the relative positions of pixels. This limitation can be addressed by using tensor-based representations of hyperspectral imagery. Tensor models, which generalize traditional vector and matrix representations, have gained significant attention in various fields, including image analysis and video processing \cite{3,4}, as well as in remote sensing imagery analysis \cite{5,6}. For example, Zhang et al. proposed a tensor discriminative locality alignment method (TDLA) for feature extraction in hyperspectral images \cite{6}, while Zhong et al. introduced LTDA, a tensorial discriminant extractor for spectral-spatial feature extraction in image classification \cite{7}.

	These studies highlight the superior representational capabilities of tensor-based multilinear algebra compared to traditional matrix algebra. Notably, Kilmer et al. introduced the t-product model \cite{8}, which defines a generalized matrix multiplication for 3D tensors based on circular convolution. This operation can be implemented efficiently using the Fourier transform. Subsequently, the t-product model has been expanded through concepts such as tubal scalars, frontal slices, and array folding/unfolding operations, helping to bridge the gap between classical linear algebra and tensor operations \cite{9,10}. These advancements have facilitated the generalization of well-established linear algebraic algorithms to the tensor domain.

	Building on the recent developments in the t-product tensor model, we propose a new tensor-based spectral-spatial feature extraction method for hyperspectral image classification. Our approach begins with the introduction of a novel tensor algebraic framework, where ``t-product" tensors are constrained to a fixed size, forming an algebraic tensor ring. This framework preserves the multi-way properties of high-order arrays while maintaining the intuitive, matrix-like structure, with ``t-product" tensors serving as the fundamental elements of our proposed tensor-vectors or tensor-matrices. Crucially, this framework is backward compatible with traditional linear algebra based on vectors and matrices.

	Leveraging this tensor algebraic framework, we extend PCA to its tensorial counterpart, Tensor PCA (TPCA), which incorporates a mechanism to capture the spatial relationships between image pixels. Our experiments, conducted on publicly available hyperspectral datasets, demonstrate that TPCA outperforms traditional PCA and other vector-based or tensor-based feature extraction methods in terms of classification accuracy.

	This paper is organized as follows. In Section 2, we present the tensor algebraic framework. In Section 3, we introduce TPCA (Tensor Principal Component Analysis) and detail its efficient implementation via the Fourier transform. Section 4 is devoted to experimental results and analysis. Finally, we conclude the paper in Section 5.

	\section{Tensor Algebra}
	In this section, we extend the ``t-product'' model \cite{8,9,10} into a novel tensor algebraic framework that is backward compatible with traditional matrix algebra. For simplicity, we focus on second-order tensors, though the proposed method can be generalized to higher-order tensors.

	Before extending the ``t-product'' model, we first constrain the tensors to a fixed size. For second-order tensors, the size is defined as $m \times n$. Consider a second-order array $X$ with entries from a field $\mathbbm{F}$. We denote its $(i,j)$-th element as $(X)_{i,j} \in \mathbbm{F}$.  
	If $X$ is a second-order array with entries from a ring $C$, we denote its $(i,j)$-th $C$-entry by $[X]_{i,j} \in C$.  
	If $X$ is a first-order array of elements from $C$, we denote its $i$-th $C$-entry by $[X]_i \in C$.
	
	The key notations and definitions related to the ``t-product tensor'' are summarized below \cite{8,9,10}.

	\begin{definition}[Tensor Addition]
		Given fixed-size second-order tensors $x_t$ and $y_t \in \mathbbm{F}^{m\times n}$, their sum $c_t = x_t + y_t \in \mathbbm{F}^{m\times n}$ is defined entry-wise by
		\[
		(c_t)_{i,j} = (x_t)_{i,j} + (y_t)_{i,j} \in \mathbbm{F}, \ \forall i,j.
		\]
	\end{definition}

	\begin{definition}[Tensor Multiplication (Circular Convolution)]
		Given second-order tensors $x_t$ and $y_t$ in $\mathbbm{F}^{m \times n}$, their product $d_t = x_t \circ y_t \in \mathbbm{F}^{m\times n}$ is defined as the two-way circular convolution:
		\[
		(d_t)_{i,j} = \sum_{k_1=1}^m \sum_{k_2=1}^n (x_t)_{k_1,k_2} (y_t)_{\operatorname{mod}(i-k_1, m)+1, \operatorname{mod}(j-k_2, n) +1}.
		\]
	\end{definition}

	The two-way circular convolution can be computed efficiently via the 2D Discrete Fourier Transform (2D-DFT) and its inverse, as shown by the following theorem:

	\begin{theorem}[Fourier Transform]
		Given $x_t, y_t \in \mathbbm{F}^{m\times n}$ and $d_t = x_t \circ y_t$, along with their 2D-DFTs $F(x_t)$, $F(y_t)$, and $F(d_t)$, we have
		\[
		\big( F(d_t) \big)_{i,j} = \big( F(x_t) \big)_{i,j} \cdot \big( F(y_t) \big)_{i,j}, \ \forall i,j.
		\]
	\end{theorem}

	Utilizing the 2D-DFT reduces the computational complexity from $(mn)^2$ scalar multiplications in the spatial domain to $mn$ independent scalar multiplications in the transformed domain, greatly accelerating convolution-based algorithms.

	We define the zero tensor $z_t$ by $(z_t)_{i,j} = 0$ for all $i,j$. The identity tensor $e_t$ is defined by $(e_t)_{1,1} = 1$ and $(e_t)_{i,j} = 0$ for all $(i,j) \neq (1,1)$.

	It can be shown that the set of $m \times n$ tensors, equipped with the addition and multiplication operations defined above, forms an algebraic ring $C$. These operations are backward compatible with those defined for the field $\mathbbm{F}$. Furthermore, by defining scalar multiplication as follows, we extend the ring $C$ into an $\mathbbm{F}$-algebra.

	\begin{definition}[Scalar Multiplication]
		Given $x_t \in C$ and $\alpha \in \mathbb{F}$, the product $d_t = \alpha \cdot x_t \in C$ is defined by
		\[
		(d_t)_{i,j} = \alpha \cdot (x_t)_{i,j} \in \mathbbm{F}, \quad \forall i,j.
		\]
	\end{definition}

	\begin{definition}[$C$-Vectors and $C$-Matrices]
		A $C$-vector is a column (or row) of $C$-entries, and a $C$-matrix is a second-order array of $C$-entries. Thus, $C$-vectors and $C$-matrices are vectors and matrices whose elements are themselves tensors in $C$.
	\end{definition}

	In our framework, operations on the entries of $C$-vectors and $C$-matrices follow the operations defined for $C$.

	\begin{definition}[$C$-Matrix Multiplication]
		Given $X_{\mathit{tm}} \in C^{M \times D}$ and $Y_{\mathit{tm}} \in C^{D \times N}$, their product $C_{\mathit{tm}} = X_{\mathit{tm}} \circ Y_{\mathit{tm}} \in C^{M \times N}$ is defined by
		\[
		[C_{\mathit{tm}}]_{i,j} = \sum_{k=1}^{D} [X_{\mathit{tm}}]_{i,k} \circ [Y_{\mathit{tm}}]_{k,j}, \quad \forall i,j.
		\]
	\end{definition}

	\begin{definition}[Identity $C$-Matrix]
		The identity $C$-matrix $I_{\mathit{tm}} \in C^{D \times D}$ is defined by
		\[
		[I_{\mathit{tm}}]_{i,j} = \begin{cases}
			e_t, & \text{if } i = j, \\
			z_t, & \text{if } i \neq j.
		\end{cases}
		\]
	\end{definition}

	\begin{definition}[Conjugate in $C$]
		For $x_t \in C \equiv \mathbbm{F}^{m\times n}$, its conjugate $x_t^{*}$ is defined by
		\[
		(x_t^*)_{i,j} = \overline{(x_t)_{\operatorname{mod}(1-i, m) + 1, \operatorname{mod}(1-j, n) + 1}} \in \mathbbm{F}, \quad \forall i,j,
		\]
		where $\overline{(\cdot)}$ denotes the complex conjugate if $\mathbbm{F} = \mathbbm{C}$ and is the identity map if $\mathbbm{F} = \mathbbm{R}$.
	\end{definition}

	\begin{definition}[Hermitian Transpose of a $C$-Matrix]
		Given $X_{\mathit{tm}} \in C^{M \times N}$, its Hermitian transpose $X_{\mathit{tm}}^{H} \in C^{N \times M}$ is defined by
		\[
		[X_{\mathit{tm}}^H]_{i,j} = [X_{\mathit{tm}}]_{j,i}^*, \quad \forall i,j.
		\]
	\end{definition}

	\begin{definition}[Unitary $C$-Matrix]
		A $C$-matrix $X_{\mathit{tm}} \in C^{D \times D}$ is unitary if
		\[
		X_{\mathit{tm}}^H \circ X_{\mathit{tm}} = X_{\mathit{tm}} \circ X_{\mathit{tm}}^H = I_{\mathit{tm}},
		\]
		where $I_{\mathit{tm}}$ is the $D \times D$ identity $C$-matrix.
	\end{definition}

	\begin{definition}[Singular Value Decomposition (SVD)]
		Every square $C$-matrix $G_{\mathit{tm}} \in C^{D \times D}$ admits a singular value decomposition of the form:
		\[
		G_{\mathit{tm}} = U_{\mathit{tm}} \circ S_{\mathit{tm}} \circ V_{\mathit{tm}}^{H},
		\]
		where $U_{\mathit{tm}}, V_{\mathit{tm}} \in C^{D \times D}$ are unitary $C$-matrices, and $S_{\mathit{tm}} \in C^{D \times D}$ is a diagonal $C$-matrix with $[S_{\mathit{tm}}]_{i,j} = z_t$ for $i \neq j$ and $[S_{\mathit{tm}}]_{i,i}^* = [S_{\mathit{tm}}]_{i,i}$ for all $i$.
	\end{definition}

	To improve computational efficiency, we extend the multi-way Discrete Fourier Transform to $C$-vectors and $C$-matrices.

	\begin{definition}[Discrete Fourier Transform of a $C$-Matrix]
		\label{def:DiscreteFourierTransform}
		For a $C$-matrix $X_{\mathit{tm}}$, its Discrete Fourier Transform $F(X_{\mathit{tm}})$ is defined entry-wise:
		\[
		[F(X_{\mathit{tm}})]_{i,j} = F([X_{\mathit{tm}}]_{i,j}), \quad \forall i,j,
		\]
		where $F([X_{\mathit{tm}}]_{i,j})$ is the multi-way discrete Fourier transform of the $(i,j)$-th $C$-entry of $X_{\mathit{tm}}$.
	\end{definition}

	Applying the discrete Fourier transform to a $C$-matrix decomposes it into a series of $\mathbbm{C}$-matrices in the transformed domain, simplifying subsequent operations.

	\begin{definition}[Slice]
		\label{def:Slice}
		Given $X_{\mathit{tm}} \in C^{M \times N}$, let $X_{\mathit{tm}}(\omega_1, \omega_2) \in \mathbbm{F}^{M \times N}$ be the slice of $X_{\mathit{tm}}$ indexed by $(\omega_1, \omega_2)$:
		\[
		(X_{\mathit{tm}}(\omega_1, \omega_2))_{i,j} = ([X_{\mathit{tm}}]_{i,j})_{\omega_1, \omega_2} \in \mathbbm{F}, \quad \forall i,j,\omega_1,\omega_2.
		\]
	\end{definition}

	Let $X_{\mathit{ftm}} \doteq F(X_{\mathit{tm}})$, and define $X_{\mathit{ftm}}(\omega_1, \omega_2) \in \mathbbm{C}^{M \times N}$ similarly:
	\[
	(X_{\mathit{ftm}}(\omega_1, \omega_2))_{i,j} = ([X_{\mathit{ftm}}]_{i,j})_{\omega_1, \omega_2}, \quad \forall i,j,\omega_1,\omega_2.
	\]

	A $C$-vector is a special case of a $C$-matrix with a single column or row; thus, slicing a $C$-vector follows the same principle.

	By Definitions \ref{def:DiscreteFourierTransform} and \ref{def:Slice}, $C$-matrix operations can be decomposed into $mn$ separate complex matrix operations in the frequency domain, enabling efficient parallel computation.

	\section{Tensor Principal Component Analysis (TPCA)}
	\subsection{Traditional PCA}

	Traditional PCA can be briefly described as follows. Given data vectors \( x_1, x_2, \dots, x_N \in \mathbbm{R}^D \) and their mean \( \bar{x} = \frac{1}{N} \sum_{k=1}^N x_k \), the covariance matrix \( G \) is defined by
	\begin{equation}
		G = \frac{1}{N-1} \sum_{k=1}^N (x_k - \bar{x})(x_k - \bar{x})^\top.
	\end{equation}
	Performing the singular value decomposition of \( G \) yields
	\begin{equation}
		G = U \Lambda U^\top,
	\end{equation}
	where \( U \in \mathbbm{R}^{D \times D} \) is an orthogonal matrix with \( U^\top U = I_{D \times D} \), and \( \Lambda = \operatorname{diag}(\sigma_1, \sigma_2, \dots, \sigma_D) \) contains the eigenvalues in non-increasing order: \( \sigma_1 \geq \sigma_2 \geq \dots \geq \sigma_D \geq 0 \).

	For a new data vector \( y \in \mathbbm{R}^D \), the PCA feature vector is given by
	\begin{equation}
		\hat{y} = U^\top (y - \bar{x}).
	\end{equation}
	To reduce the dimension from \( D \) to \( d \) (\( d < D \)), we keep only the first \( d \) components of \(\hat{y}\) corresponding to the largest eigenvalues \(\sigma_1, \sigma_2, \dots, \sigma_d\).

	\subsection{Tensor PCA (TPCA)}

	TPCA extends traditional PCA to tensorial data, allowing us to incorporate spatial structures within the data representation. Let \( X_{\mathit{tv},1}, X_{\mathit{tv},2}, \dots, X_{\mathit{tv},N} \in C^{D} \equiv \mathbbm{F}^{m\times n \times D} \) be a set of high-order samples. Define the mean $C$-vector as
	\begin{equation}
		\bar{X}_{\mathit{tv}} = \frac{1}{N} \sum_{k=1}^N X_{\mathit{tv},k}.
	\end{equation}
	The covariance $C$-matrix \( G_{\mathit{tm}} \) is given by
	\begin{equation}
		G_{\mathit{tm}} = \frac{1}{N-1} \sum_{k=1}^N (X_{\mathit{tv},k} - \bar{X}_{\mathit{tv}}) \circ (X_{\mathit{tv},k} - \bar{X}_{\mathit{tv}})^{H},
	\end{equation}
	where \( \circ \) denotes the $C$-matrix multiplication defined in Section 2.

	Performing the tensorial singular value decomposition (TSVD) of \( G_{\mathit{tm}} \in C^{D \times D} \) gives
	\begin{equation}
		G_{\mathit{tm}} = U_{\mathit{tm}} \circ S_{\mathit{tm}} \circ V_{\mathit{tm}}^{H},
		\label{equation:G-TSVD}
	\end{equation}
	where \( U_{\mathit{tm}} \) and \( V_{\mathit{tm}} \) are unitary $C$-matrices, and \( S_{\mathit{tm}} \) is a diagonal $C$-matrix containing the singular values in the $C$-algebraic framework.

	For a new $C$-vector sample \( Y_{\mathit{tv}} \in C^{D} \), the TPCA feature $C$-vector is obtained by
	\[
	\hat{Y}_{\mathit{tv}} = U_{\mathit{tm}}^H \circ (Y_{\mathit{tv}} - \bar{X}_{\mathit{tv}}).
	\]

	To convert the $C$-vector feature \( \hat{Y}_{\mathit{tv}} \) into a traditional $\mathbbm{F}$-vector suitable for standard algorithms, we define a mapping \( \delta: C^{D} \equiv \mathbbm{F}^{m\times n\times D} \rightarrow \mathbbm{F}^{D} \):
	\begin{equation}
		\delta(\hat{Y}_{\mathit{tv}}) = \frac{1}{mn} \sum_{\omega_1=1}^m \sum_{\omega_2=1}^n \hat{Y}_{\mathit{tv}}(\omega_1, \omega_2).
		\label{equation:delta}
	\end{equation}

	To map the $C$-vector feature \(\hat{Y}_{\mathit{tv}}\) into a traditional $\mathbbm{F}$-vector suitable for conventional algorithms, define the mapping \(\delta: C^{D} \equiv \mathbbm{F}^{m\times n\times D} \to \mathbbm{F}^{D}\) as
	\begin{equation}
		\delta(\hat{Y}_{\mathit{tv}}) = \frac{1}{mn} \sum_{\omega_1=1}^m \sum_{\omega_2=1}^n \hat{Y}_{\mathit{tv}}(\omega_1, \omega_2).
		\label{equation:delta}
	\end{equation}

	Thus, the traditional TPCA feature vector is
	\[
	\hat{y} = \delta\big(U_{\mathit{tm}}^H \circ (Y_{\mathit{tv}} - \bar{X}_{\mathit{tv}})\big) \in \mathbbm{F}^{D}.
	\]
	To reduce the dimension of \(\hat{y}\) from \(D\) to \(d\), we simply discard the last \((D-d)\) entries.
	
	The TPCA procedure is summarized in Algorithm \ref{alg:TPCA}.

	\begin{algorithm}[H]
		\caption{Tensor Principal Component Analysis (TPCA)}
		\textbf{Input:} A query $C$-vector \( Y_{\mathit{tv}} \in C^{D} \equiv \mathbbm{F}^{m\times n \times D} \) and \( N \) training $C$-vectors \( X_{\mathit{tv}, 1}, \dots, X_{\mathit{tv}, N} \in C^{D} \equiv \mathbbm{F}^{m\times n \times D} \).\\
		\textbf{Output:} TPCA feature $\mathbbm{F}$-vector \( y \in \mathbbm{F}^D \).
		\begin{algorithmic}[1]
			\State Compute the mean 
			$\bar{X}_{\mathit{tv}} = \frac{1}{N} \sum_{i=1}^N X_{\mathit{tv}, i}.$
			\State Compute the covariance $C$-matrix
			$
			G_{\mathit{tm}} = \frac{1}{N-1} \sum_{i=1}^N ( X_{\mathit{tv}, i} - \bar{X}_{\mathit{tv}} ) \circ ( X_{\mathit{tv}, i} - \bar{X}_{\mathit{tv}} )^{H}.
			$
			\State Perform the TSVD of \( G_{\mathit{tm}} \):
			$
			G_{\mathit{tm}} = U_{\mathit{tm}} \circ S_{\mathit{tm}} \circ V_{\mathit{tm}}^H.
			$
			\State Compute the $C$-vector feature:
			$
			\hat{Y}_{\mathit{tv}} = U_{\mathit{tm}}^{H} \circ (Y_{\mathit{tv}} - \bar{X}_{\mathit{tv}}).
			$
			\State Map \( \hat{Y}_{\mathit{tv}} \) to a traditional $\mathbbm{F}$-vector:
			$
			y = \delta(\hat{Y}_{\mathit{tv}}) \in \mathbbm{F}^{D},
			$
			where \(\delta(\cdot)\) is defined by Equation \eqref{equation:delta}.
			\State \textbf{return} \( y \).
		\end{algorithmic}
	\label{alg:TPCA}
	\end{algorithm}

	\subsection{Fast TPCA Using the Fourier Transform}

	The Fourier transform enables efficient computation of TPCA by decomposing $C$-matrix operations into parallelizable 
	$\mathbbm{F}$-matrix operations in the frequency domain. This is achieved through slicing and applying the discrete Fourier transform (DFT). For example, the fast implementation of Equation \eqref{equation:G-TSVD} is detailed in Algorithm \ref{alg:FourierTPCA}.

	\begin{algorithm}[H]
		\caption{Fast Implementation of the TSVD of $G_{\mathit{tm}}$ in Equation \eqref{equation:G-TSVD}}
		\label{alg:FourierTPCA}
		\textbf{Input:} Covariance $C$-matrix \( G_{\mathit{tm}} \in C^{D \times D} \).\\
		\textbf{Output:} $C$-matrices \( U_{\mathit{tm}}, S_{\mathit{tm}}, V_{\mathit{tm}} \).
		\begin{algorithmic}[1]
			\State Compute the discrete Fourier transform \( G_{\mathit{ftm}} = F(G_{\mathit{tm}}) \).
			\ForAll{frequency components \( (\omega_1, \omega_2) \in \{1, \dots, m\} \times \{1, \dots, n\} \)}
			\State Compute the $\mathbbm{F}$-matrix SVD of the slice:
			$
			G_{\mathit{ftm}}(\omega_1, \omega_2) = U \cdot S \cdot V^H.
			$
			\State Set \( U_{\mathit{ftm}}(\omega_1, \omega_2) = U \), \( S_{\mathit{ftm}}(\omega_1, \omega_2) = S \), 
			\( V_{\mathit{ftm}}(\omega_1, \omega_2) = V \).
			\EndFor
			\State Compute the inverse Fourier transforms:
			$
			U_{\mathit{tm}} = F^{-1}(U_{\mathit{ftm}}), \quad S_{\mathit{tm}} = F^{-1}(S_{\mathit{ftm}}), \quad V_{\mathit{tm}} = F^{-1}(V_{\mathit{ftm}}).
			$
			\State \textbf{return} \( U_{\mathit{tm}}, S_{\mathit{tm}}, V_{\mathit{tm}} \).
		\end{algorithmic}
	\end{algorithm}

	The entire TPCA workflow can also be accelerated using discrete Fourier transforms. Algorithm \ref{alg:FastTPCA} summarizes this process.

	\begin{algorithm}[H]
		\caption{Fast Implementation of TPCA}
		\label{alg:FastTPCA}
		\textbf{Input:} Query tensor \( Y_{\mathit{tv}} \in C^{D} \) and \( N \) training tensors \( X_{\mathit{tv}, 1}, \dots, X_{\mathit{tv}, N} \in C^{D} \).\\
		\textbf{Output:} TPCA feature $\mathbbm{F}$-vector \( y \in \mathbbm{F}^D \).
		\begin{algorithmic}[1]
			\State Compute \( \bar{X}_{\mathit{tv}} = \frac{1}{N} \sum_{i=1}^N X_{\mathit{tv}, i} \).
			\State Compute the Fourier transforms \( X_{\mathit{ftv},i} = F( X_{\mathit{tv}, i} - \bar{X}_{\mathit{tv}} ) \) and 
			\( Y_{\mathit{ftv}} = F(Y_{\mathit{tv}} - \bar{X}_{\mathit{tv}}) \), for \( i = 1, \dots, N \).
			\ForAll{frequency components \( (\omega_1, \omega_2) \in \{1, \dots, m\} \times \{1, \dots, n\} \)}
			\State Compute 
			$
			G_{\mathit{ftm}}(\omega_1, \omega_2) = \frac{1}{N-1} \sum_{i=1}^N X_{\mathit{ftv},i}(\omega_1, \omega_2) \cdot X_{\mathit{ftv},i}(\omega_1, \omega_2)^H.
			$
			\State Compute the $\mathbbm{F}$-matrix SVD:
			$
			G_{\mathit{ftm}}(\omega_1, \omega_2) = U \cdot S \cdot V^H.
			$
			\State Compute 
			$
			\hat{Y}_{\mathit{ftv}}(\omega_1, \omega_2) = U^H \cdot Y_{\mathit{ftv}}(\omega_1, \omega_2).
			$
			\EndFor
			\State Compute the inverse Fourier transform \( \hat{Y}_{\mathit{tv}} = F^{-1}(\hat{Y}_{\mathit{ftv}}) \).
			\State Map \( \hat{Y}_{\mathit{tv}} \) to a traditional $\mathbbm{F}$-vector \( y = \delta(\hat{Y}_{\mathit{tv}}) \).
			\State \textbf{return} \( y \).
		\end{algorithmic}
	\end{algorithm}

	\subsection{Computational Complexity}
	
	The proposed acceleration scheme decomposes a tensorial operation (such as TPCA or TSVD) into \( mn \) separate, traditional (non-tensorial) operations performed in the Fourier domain. Here, \( mn \) is the number of frequency components, equal to the total number of slices.
	
	Assuming the cost of performing a traditional operation in the Fourier domain is proportional to one floating-point operation (flop), the computational cost of the corresponding tensorial operation is proportional to \( mn \) flops. Neglecting the overhead of the Fourier transform, the computational complexity of TPCA or TSVD thus scales linearly with \( mn \), i.e., it is \( \mathcal{O}(mn) \).

	\section{Experiments}
	
	We use two publicly available hyperspectral image datasets for our experiments: the Indian Pines scene and the Pavia University scene. The Indian Pines dataset contains hyperspectral images with a spatial resolution of $145 \times 145$ pixels over 200 spectral bands. Ground-truth labels are provided for 16 land-cover classes. The Pavia University dataset has hyperspectral images with a spatial resolution of $610 \times 340$ pixels across 103 spectral bands, with ground-truth labels for 9 land-cover classes. Both datasets are widely recognized benchmarks in hyperspectral image analysis.

	\subsection{Tensorization}

	To facilitate tensor-based analysis of hyperspectral data, we apply a tensorization process that integrates both spatial and spectral information. Consider a hyperspectral image \(\mathcal{H} \in \mathbbm{R}^{M \times N \times D}\), where \(M\) and \(N\) are the spatial dimensions, and \(D\) is the number of spectral bands. Let \(Z_{i,j} \in \mathbbm{R}^D\) denote the spectral vector at the pixel \((i,j)\).

	To enable tensorial analysis of hyperspectral data, we apply a tensorization process that integrates spatial and spectral information. Given a hyperspectral image represented as \( \mathcal{H} \in \mathbbm{R}^{M \times N \times D} \), where \( M \) and \( N \) are spatial dimensions and \( D \) is the number of spectral bands, let \( Z_{i,j} \in \mathbbm{R}^D \) denote the spectral vector at pixel \((i,j)\). For each pixel, we construct a $C$-vector \( Z_{\mathit{tv},i,j} \in C^{D} \equiv \mathbbm{R}^{3 \times 3 \times D} \) by capturing its local spatial neighborhood in all spectral channels.

	From the hyperspectral image \(\mathcal{H}\), we randomly select \(N\) pixels, ensuring a representative coverage across the image. Each selected pixel is tensorized as described above, producing \(N\) samples \( Z_{\mathit{tv},i,j} \in C^{D} \equiv \mathbbm{R}^{3 \times 3 \times D} \). These tensorized samples serve as the input data for the Tensor Principal Component Analysis (TPCA) procedure detailed in Algorithm~\ref{alg:TPCA}.

	This tensorization approach preserves the spatial and spectral characteristics of hyperspectral pixels, thereby enabling more effective feature extraction and dimensionality reduction through TPCA.

	\subsection{Experimental Results}
	
	The performance of our method is evaluated using Overall Accuracy (OA) and the $\kappa$ coefficient, two widely used metrics in classification tasks. OA measures the proportion of correctly classified pixels, while the $\kappa$ coefficient quantifies the agreement between predicted and actual labels, accounting for chance agreement \cite{11}. Higher values of both metrics indicate better classification results.

	We compare five feature extractors with three classifiers. The classifiers include Nearest Neighbour (NN), Support Vector Machine (SVM), and Random Forest (RF). For SVM, a Gaussian RBF kernel is employed, with the Gaussian parameter $\sigma$ and regularization parameter $C$ optimized using 5-fold cross-validation. The parameters are chosen from $\sigma \in \{2^i\}_{i=-15}^{10}$ and $C \in \{2^i\}_{i=-5}^{15}$.

	The feature extractors include two classical vector-based methods, PCA and LDA, and three state-of-the-art tensor-based approaches: TDLA \cite{6}, LTDA \cite{7}, and TPCA (our method). Additionally, the original raw vector representation is used as a baseline for comparison. For PCA, LDA, and TPCA, we evaluate the classification performance across feature dimensions \( D \in \{5, 10, \dots, D_{\mathrm{max}}\} \), where \( D_{\mathrm{max}} = 200 \) for the Indian Pines dataset and \( D_{\mathrm{max}} = 100 \) for the Pavia University dataset. The highest classification accuracy across the range of \( D \) values is reported for each method.

	In our experiments, 10\% of the pixels from the hyperspectral images are uniformly randomly selected as training samples, while the remaining pixels are used as query samples. The experiments are repeated independently 10 times to ensure statistical reliability, and the average OA and $\kappa$ values are reported.
	
	This evaluation framework ensures a fair and robust comparison of feature extractors and classifiers, highlighting the effectiveness of our proposed TPCA method.

	\medskip

\begin{figure}[H]
		\centering
		\begin{tabular}{
				>{\centering\arraybackslash}m{0.125\textwidth}  
				>{\centering\arraybackslash}m{0.125\textwidth} 
				>{\centering\arraybackslash}m{0.125\textwidth}  
				>{\centering\arraybackslash}m{0.125\textwidth}  
				>{\centering\arraybackslash}m{0.125\textwidth}  
				>{\centering\arraybackslash}m{0.125\textwidth}  
				>{\centering\arraybackslash}m{0.125\textwidth}            
				>{\centering\arraybackslash}m{0.125\textwidth} }
			\multicolumn{8}{l}{     
				\includegraphics[width=0.125\textwidth]{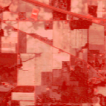}
				\includegraphics[width=0.125\textwidth]{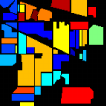}
				\includegraphics[width=0.125\textwidth]{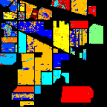}
				\includegraphics[width=0.125\textwidth]{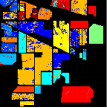}
				\includegraphics[width=0.125\textwidth]{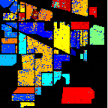}
				\includegraphics[width=0.125\textwidth]{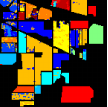}
				\includegraphics[width=0.125\textwidth]{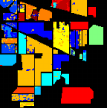}
				\includegraphics[width=0.125\textwidth]{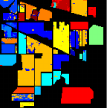}
			} \\
			\multicolumn{8}{l}{     
				\hspace{3em}(a)
				\hspace{4.5em}(b)
				\hspace{4.5em}(c)
				\hspace{4.9em}(d)
				\hspace{4.8em}(e)
				\hspace{4.9em}(f)
				\hspace{4.9em}(g)
				\hspace{4.9em}(h)
			}     
		\end{tabular}  
		\caption{Classification maps generated using RF with different types of features on the Indian Pines dataset. (a) Indian Pines scene, (b) Ground-truth, (c) Original, (d) PCA, (e) LDA, (f) TDLA, (g) LTDA, (h) TPCA.}     
		\label{fig:002}
	\end{figure}

	The quantitative comparison of the classification results is presented in Table \ref{table:classification}. The third and fourth columns of Table \ref{table:classification} show the OA and $\kappa$ obtained on the Indian Pines dataset. It is evident that the tensorial features (TDLA, LTDA, and TPCA) outperform the vectorial features (PCA and LDA). Furthermore, among all extractors, TPCA consistently yields the highest OA and the largest $\kappa$, surpassing its tensorial counterparts TDLA and LTDA. Specifically, the OA obtained by TPCA is approximately 6\%–11\% higher than that obtained by PCA. For instance, using RF, TPCA achieves an accuracy of 91.01\% compared to the baseline 76.78\% and PCA's 79.78\%. The Indian Pines dataset, along with the ground-truth and classification maps with the highest accuracies for different experimental settings, are shown in Figure \ref{fig:002}. It is clear that TPCA provides the best classification results.

\begin{figure}[H]
		\centering
		\begin{tabular}{
				>{\centering\arraybackslash}m{0.125\textwidth}  
				>{\centering\arraybackslash}m{0.125\textwidth} 
				>{\centering\arraybackslash}m{0.125\textwidth}  
				>{\centering\arraybackslash}m{0.125\textwidth}  
				>{\centering\arraybackslash}m{0.125\textwidth}  
				>{\centering\arraybackslash}m{0.125\textwidth}  
				>{\centering\arraybackslash}m{0.125\textwidth}            
				>{\centering\arraybackslash}m{0.125\textwidth} }
			\multicolumn{8}{l}{     
				\includegraphics[width=0.125\textwidth]{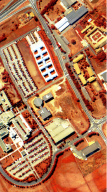}
				\includegraphics[width=0.125\textwidth]{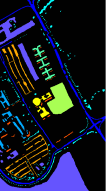}
				\includegraphics[width=0.125\textwidth]{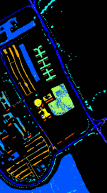}
				\includegraphics[width=0.125\textwidth]{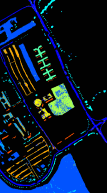}
				\includegraphics[width=0.125\textwidth]{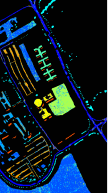}
				\includegraphics[width=0.125\textwidth]{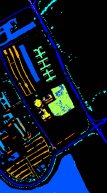}
				\includegraphics[width=0.125\textwidth]{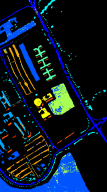}
				\includegraphics[width=0.125\textwidth]{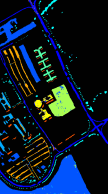}
			} \\
			\multicolumn{8}{l}{     
				\hspace{3em}(a)
				\hspace{4.5em}(b)
				\hspace{4.5em}(c)
				\hspace{4.9em}(d)
				\hspace{4.8em}(e)
				\hspace{4.9em}(f)
				\hspace{4.9em}(g)
				\hspace{4.9em}(h)
			}     
		\end{tabular}  
		\caption{Classification maps generated using RF with different types of features on the Pavia University dataset. (a) Pavia University scene, (b) Ground-truth, (c) Original, (d) PCA, (e) LDA, (f) TDLA, (g) LTDA, (h) TPCA.}      
		\label{fig:003}
	\end{figure}

	Since PCA and TPCA have similar structures, we compare their performances by plotting the OA curves of PCA and TPCA across varying feature dimensions. These curves, obtained using classifiers NN, SVM, and RF on the Indian Pines dataset, are shown in Figure \ref{fig:003}. The figure demonstrates that, regardless of the classifier and feature dimension chosen, TPCA consistently outperforms PCA.

\begin{figure}[H]
		\centering
		\begin{tabular}{>{\centering\arraybackslash}m{0.33\textwidth}  >{\centering\arraybackslash}m{0.33\textwidth}  >{\centering\arraybackslash}m{0.33\textwidth} }
			\includegraphics[width=0.33\textwidth]{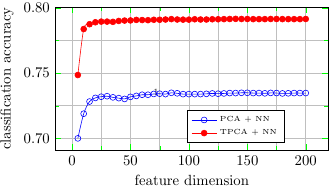} & 
			\includegraphics[width=0.33\textwidth]{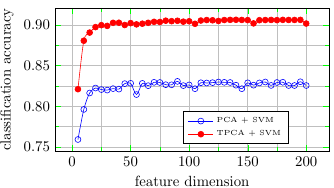} & 
			\includegraphics[width=0.33\textwidth]{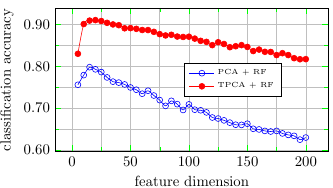} \\ 
			~~~~~~~(a) & ~~~~~~~(b) & ~~~~~~~(c)         
		\end{tabular}
		\caption{Classification accuracy curves for the Indian Pines dataset using PCA and TPCA combined with NN, SVM, and RF classifiers. (a) NN, (b) SVM, and (c) RF.}
		\label{fig:004}
	\end{figure}

\begin{figure}[H]
		\centering
		\begin{tabular}{>{\centering\arraybackslash}m{0.33\textwidth}  >{\centering\arraybackslash}m{0.33\textwidth}  >{\centering\arraybackslash}m{0.33\textwidth} }
			\includegraphics[width=0.33\textwidth]{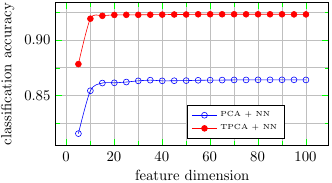} & 
			\includegraphics[width=0.33\textwidth]{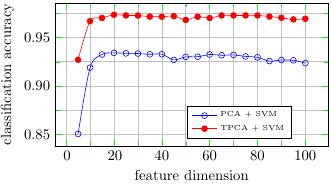} & 
			\includegraphics[width=0.33\textwidth]{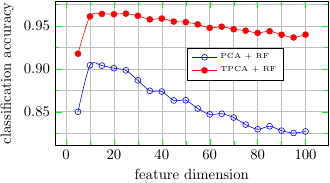} \\ 
			~~~~~~~(a) & ~~~~~~~(b) & ~~~~~~~(c)         
		\end{tabular}
		\caption{Classification accuracy curves for the Pavia University dataset using PCA and TPCA combined with NN, SVM, and RF classifiers. (a) NN, (b) SVM, and (c) RF.}
		\label{fig:005}
	\end{figure}

	The classification results obtained using various classifiers and feature extractors on the Pavia University dataset are presented in the last two columns of Table \ref{table:classification}. The visual classification maps generated by RF with different extractors are shown in Figure \ref{fig:004}. Additionally, the OA curves of PCA and TPCA for varying feature dimensions, obtained using classifiers NN, SVM, and RF on the Pavia University dataset, are depicted in Figure \ref{fig:005}. From Table \ref{table:classification}, Figure \ref{fig:004}, and Figure \ref{fig:005}, we observe that TPCA consistently outperforms its rivals, supporting our claims about its effectiveness.

	The classification results obtained using various classifiers and feature extractors on the Pavia University dataset are presented in the last two columns of Table \ref{table:classification}. The visual classification maps generated by RF with different extractors are shown in Figure \ref{fig:004}. Additionally, the OA curves of PCA and TPCA for varying feature dimensions, obtained using classifiers NN, SVM, and RF on the Pavia University dataset, are depicted in Figure \ref{fig:005}. From Table \ref{table:classification}, Figure \ref{fig:004}, and Figure \ref{fig:005}, we observe that TPCA consistently outperforms its rivals, supporting our claims about its effectiveness.

	\begin{table}[htbp]
		\centering
		\caption{Classification accuracy comparison obtained on the Indian Pines and Pavia University datasets}
		\label{table:classification}
		\small
		\begin{tabular}{c|c|cc|cc}
			\hline
			
			\hline
			
			\hline
			\multirow{2}{*}{Classifier} & \multirow{2}{*}{Extractor} & \multicolumn{2}{c|}{Indian Pines} & \multicolumn{2}{c}{Pavia University} \\
			\cline{3-6}
			&  & OA & $\kappa$ & OA & $\kappa$ \\
			\hline
			\hline
			\multirow{6}{*}{NN} & Original & 73.43 & 0.6972 & 86.35 & 0.8170 \\
			& PCA & 73.49 & 0.6977 & 86.40 & 0.8178 \\
			& LDA & 71.54 & 0.6642 & 85.04 & 0.8037 \\
			& TDLA \cite{6} & 74.21 & 0.7209 & 89.27 & 0.8568 \\
			& LTDA \cite{7} & 75.14 & 0.7216 & 90.48 & 0.8718 \\
			& TPCA [ours] & \textbf{79.15} & \textbf{0.7624} & \textbf{92.35} & \textbf{0.8979} \\
			\hline
			\multirow{6}{*}{SVM} & Original & 82.95 & 0.8053 & 93.60 & 0.9143 \\
			& PCA & 83.06 & 0.8065 & 93.41 & 0.9123 \\
			& LDA & 79.53 & 0.7673 & 89.76 & 0.8606 \\
			& TDLA \cite{6} & 83.51 & 0.8168 & 96.14 & 0.9498 \\
			& LTDA \cite{7} & 85.68 & 0.8372 & 94.91 & 0.9324 \\
			& TPCA [ours] & \textbf{90.62} & \textbf{0.8930} & \textbf{97.34} & \textbf{0.9648} \\
			\hline
			\multirow{6}{*}{RF} & Original & 76.78 & 0.7315 & 89.79 & 0.8624 \\
			& PCA & 79.78 & 0.7667 & 90.42 & 0.8712 \\
			& LDA & 76.59 & 0.7353 & 87.74 & 0.8395 \\
			& TDLA \cite{6} & 84.96 & 0.8237 & 94.32 & 0.9245 \\
			& LTDA \cite{7} & 86.57 & 0.8475 & 93.03 & 0.9178 \\
			& TPCA [ours] & \textbf{91.01} & \textbf{0.8969} & \textbf{96.44} & \textbf{0.9526} \\
			\hline
			
			\hline
			
			\hline
		\end{tabular}
	\end{table}

	\section{Conclusion}
	
	We propose a novel tensor-based feature extractor called TPCA (Tensor Principal Component Analysis) for hyperspectral image classification. First, we introduce a new tensor matrix algebraic framework that integrates the recently developed t-product model, based on circular convolution, with traditional matrix algebra. Utilizing this framework, we extend the conventional PCA algorithm to its tensorial variant, TPCA. To enhance computational efficiency, we also present a fast TPCA implementation where calculations are performed in the Fourier domain.
	
	Employing a tensorization scheme that incorporates the local spatial neighborhood of each pixel, each sample is represented by a tensorial vector whose entries are all second-order tensors. This allows TPCA to effectively extract both spectral and spatial information from hyperspectral images. Additionally, to facilitate the use of TPCA with traditional vector-based classifiers, we design a straightforward and effective method to convert TPCA's output tensor vectors into traditional vectors.
	
	Experiments conducted on two publicly available benchmark hyperspectral datasets demonstrate that TPCA outperforms its competitors, including PCA, LDA, TDLA, and LTDA, in terms of classification accuracy.
	
	\section*{Acknowledgements}
	
	This work is funded by the National Natural Science Foundation of China (No. U1404607), the Open Foundation Program of the Shaanxi Provincial Key Laboratory of Speech and Image Information Processing (No. SJ2013001), the High-end Foreign Experts Recruitment Program (No. GDW20134100119), and the Key Science and Technology Program of Henan (No. 142102210557).
	
	In this work, the development of the algebraic framework for tensorial matrices and TPCA was primarily carried out by Liang Liao and Stephen Maybank. The experimental evaluation on hyperspectral data was mainly performed by Yuemei Ren, with substantial additional contributions from Liang Liao.

\section*{Citation}

	This paper is a reconstructed and refined version of the original publication in 2017. Please cite this paper as follows:
	
	\medskip
	Ren, Yuemei, Liang Liao, Stephen John Maybank, Yanning Zhang, and Xin Liu. ``Hyperspectral Image Spectral-Spatial Feature Extraction via Tensor Principal Component Analysis.'' \textit{IEEE Geoscience and Remote Sensing Letters} 14, no.~9 (2017): 1431--1435.

\end{document}